\pdfoutput=1

\documentclass[11pt]{article}

\usepackage[preprint]{acl}

\usepackage{times}
\usepackage{latexsym}

\usepackage[T1]{fontenc}

\usepackage[utf8]{inputenc}

\usepackage{microtype}

\usepackage{inconsolata}

\usepackage{xspace}
\usepackage{inconsolata}
\usepackage{booktabs}
\usepackage{comment}
\usepackage{graphicx} 
\usepackage{tcolorbox} 
\usepackage{framed}
\usepackage{CJKutf8} 
\usepackage{multirow}
\usepackage{url}
\usepackage{svg}
\usepackage{soul}
\usepackage{fontawesome5}
\usepackage{xltabular} 

\usepackage{amsthm}

\usepackage{tikz}
\usepackage{forest}
\usetikzlibrary{trees,positioning,shapes,shadows,arrows.meta}
\usepackage{amssymb}
\usepackage{adjustbox}
\usepackage{rotating}
\usepackage{array}
\usepackage{makecell}

\newcommand\blfootnote[1]{%
  \begingroup
  \renewcommand\thefootnote{}\footnote{#1}%
  \addtocounter{footnote}{-1}%
  \endgroup
}

\title{The Potential and Challenges of Evaluating Attitudes, Opinions, and Values in Large Language Models}

\vspace{8pt}

\author{
 \textbf{Bolei Ma\textsuperscript{$\ast$ \includegraphics[scale=0.014]{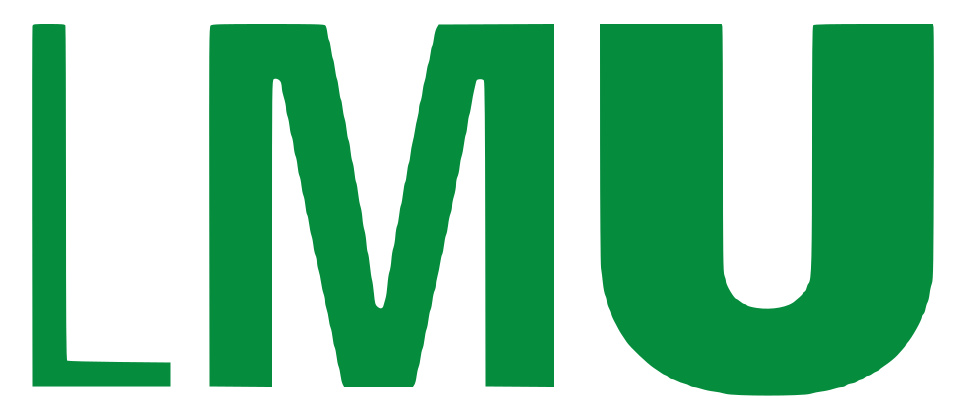}, \includegraphics[scale=0.023]{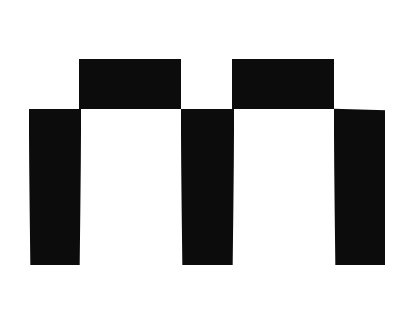}}}~~~
 \textbf{Xinpeng Wang\textsuperscript{$\ast$ \includegraphics[scale=0.014]{figs/lmu.png}, \includegraphics[scale=0.023]{figs/mcml_m.png}}}~~~
 \textbf{Tiancheng Hu\textsuperscript{\includegraphics[scale=0.023]{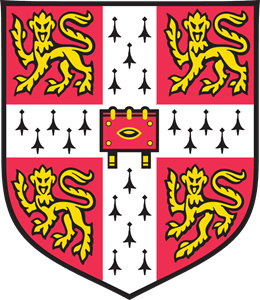}}}~~~
  \textbf{Anna-Carolina Haensch\textsuperscript{\includegraphics[scale=0.014]{figs/lmu.png}, \includegraphics[scale=0.006]{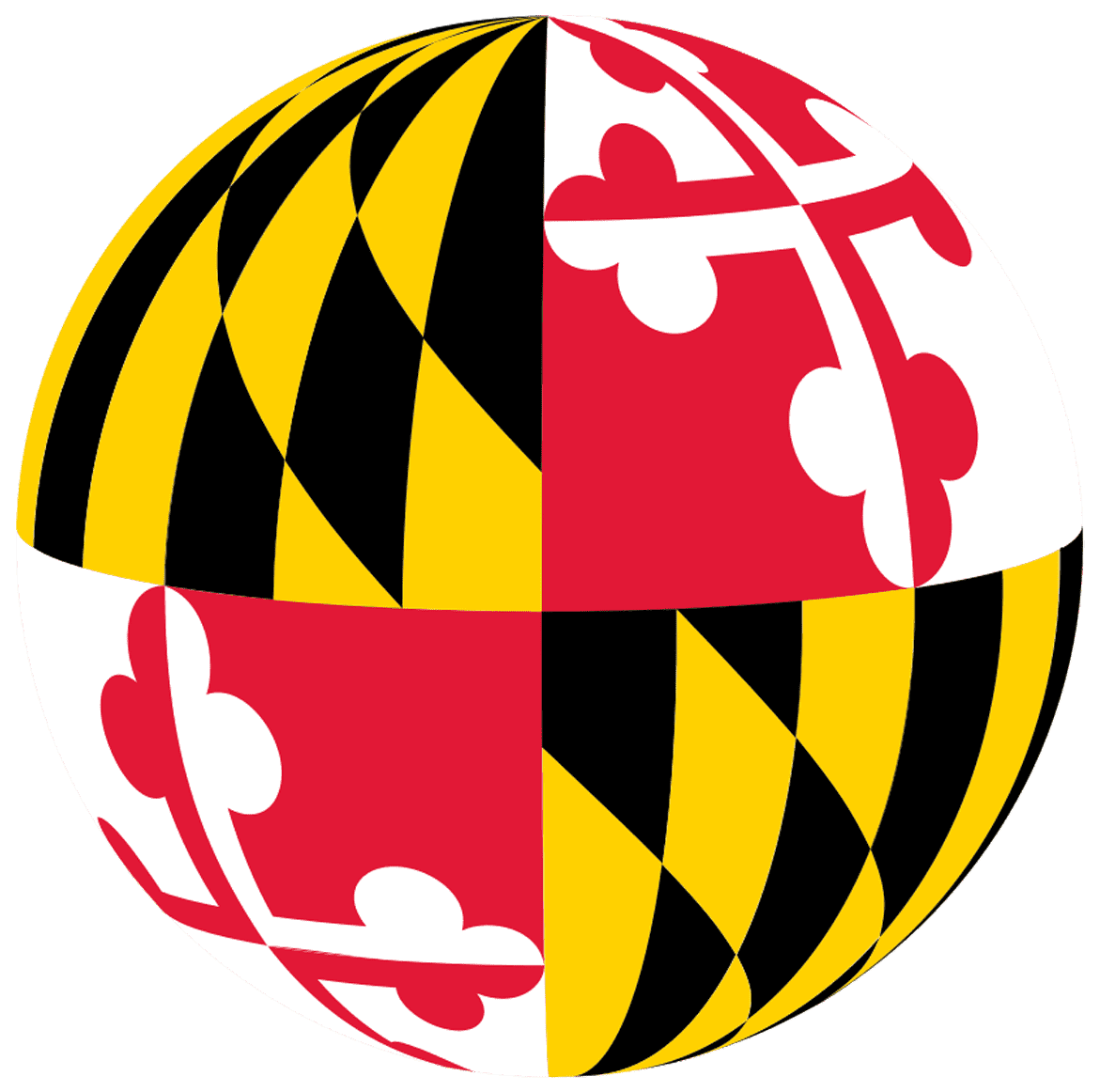}}}~~~
\\
  \textbf{Michael A. Hedderich\textsuperscript{\includegraphics[scale=0.014]{figs/lmu.png}, \includegraphics[scale=0.023]{figs/mcml_m.png}}}~~~
 \textbf{Barbara Plank\textsuperscript{\includegraphics[scale=0.014]{figs/lmu.png}, \includegraphics[scale=0.023]{figs/mcml_m.png}, \includegraphics[scale=0.2]{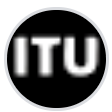}}}~~~
 \textbf{Frauke Kreuter\textsuperscript{\includegraphics[scale=0.014]{figs/lmu.png}, \includegraphics[scale=0.023]{figs/mcml_m.png}, \includegraphics[scale=0.006]{figs/University-of-Maryland-Logo.png}}}
\vspace{7.5pt}
\\
 \textsuperscript{\includegraphics[scale=0.014]{figs/lmu.png}}LMU Munich~~~
 \textsuperscript{\includegraphics[scale=0.023]{figs/mcml_m.png}}Munich Center for Machine Learning~~~
 \textsuperscript{\includegraphics[scale=0.023]{figs/university-of-cambridge-logo.png}}University of Cambridge~~~\\
 \textsuperscript{\includegraphics[scale=0.006]{figs/University-of-Maryland-Logo.png}}University of Maryland, College Park~~~
 \textsuperscript{\includegraphics[scale=0.2]{figs/itu_round.png}}ITU Copenhagen
\vspace{7.5pt}
\\
\texttt{bolei.ma@lmu.de,~xinpeng@cis.lmu.de}
}

\begin{document}
\maketitle
\begin{abstract}
Recent advances in Large Language Models (LLMs) have sparked wide interest in validating and comprehending the human-like cognitive-behavioral traits LLMs may capture and convey. 
These cognitive-behavioral traits include typically \textit{Attitudes, Opinions, Values} (AOVs).
However, measuring AOVs embedded within LLMs remains opaque, and different evaluation methods may yield different results. 
This has led to a lack of clarity on how different studies are related to each other and how they can be interpreted.
This paper aims to bridge this gap by providing a comprehensive overview of recent works on the evaluation of AOVs in LLMs. Moreover, we survey related approaches in different stages of the evaluation pipeline in these works. By doing so, we address the potential and challenges with respect to understanding the model, human-AI alignment, and downstream application in social sciences. 
Finally, we provide practical insights into evaluation methods, model enhancement, and interdisciplinary collaboration, thereby contributing to the evolving landscape of evaluating AOVs in LLMs.\blfootnote{$^\ast$Lead authors.}
\end{abstract}

\section{Introduction}
Recent years have witnessed a remarkable improvement in the development of Large Language Models (LLMs), holding the promise of boosting various domains, from computer sciences to social sciences and beyond \cite{ziems-etal-2024-large}. Amid the excitement surrounding their capabilities, when we take a human-centric perspective, an important question lies: How well do these LLMs capture and convey human cognitive-behavioral traits?

By drawing upon traditional theories from the social sciences (such as \citealp{Katz1960, rokeach1968beliefs, ajzen1988attitudes, bergman1998}), we consider human cognitive-behavioral traits, in our case specifically \textbf{Attitudes, Opinions, Values (AOVs)}, as fundamental components of human cognition, shaping our perceptions, decisions, and interactions. 
By examining whether and how LLM outputs reflect AOVs, and comparing these AOVs to those of humans, we can gain deeper insights into the models' capacity to function as autonomous agents mirroring human AOVs. The AOVs in LLMs also impact users in downstream applications, such as writing assistants \cite{Jakesch2023}, and affect decision-making processes and perceptions 
\cite{eigner2024determinants}.

In recent studies, survey questionnaires that were originally used to estimate public opinions in the social sciences are now being popularly utilized to evaluate the opinions of LLMs and subsequently to study the alignment with human opinions \cite[][\emph{inter alia}]{santurkar2023whose, hwang-etal-2023-aligning, kim2024aiaugmented}. 
At the same time, the wide range of evaluation methods used to assess LLM responses has led to inconsistent outcomes, complicating reliable assessment of the models. For instance, \citet{rottger2024political} demonstrate that different prompting methods lead to different results; \citet{wang2024my} show that output extraction methods can yield misaligned results. 
However, this variability in evaluation methods has sometimes been overlooked in real usecases---posing risks of missing subtleties in LLM performance, yielding incomplete or biased assessments. This oversight raises questions about the model's true capabilities and its alignment with human opinions.

Motivated by the rising interest in studying the human-like traits of LLMs, in this paper, we present the first survey on the evaluation of AOVs in LLMs. 
Before moving into the details, we first position our survey in the context of other relevant surveys and then show the framework of our survey.

\paragraph{Related Survey Papers.}

While there are no survey papers specifically on AOVs in LLMs, some existing works have covered related questions. On opinions, \citet{simmons2023large} review frameworks for using LLMs to model subpopulations and measure public opinions. Similarly, \citet{JANSEN2023} explore the use of LLMs in public opinion research, noting their potential to enhance survey methods. On values, \citet{vida-etal-2023-values} survey moral NLP, emphasizing the need for more rigorous discussions on ethical concepts. \citet{hershcovich-etal-2022-challenges} examine NLP in cross-cultural contexts, urging the preservation of cultural values. Additionally, recent surveys on understanding ``culture'' in NLP \cite{liu2024culturally} and measuring ``cultures'' in LLMs \cite{adilazuarda2024measuring} highlight the importance of culturally aware techniques. These works have mainly explored topics like improving public opinion research with LLMs or studying the cultural and moral aspects in NLP. However, there has been a lack of dedicated studies focusing on AOVs and especially on evaluating the AOVs within LLMs.

\paragraph{Our Survey Paper.}
Since LLMs are pretrained on vast amounts of human data, it is reasonable to hypothesize that LLMs can reflect the AOVs embedded in the data \cite{durmus2024towards}. 
But, for that to scale, we will need definitions of the terms AOVs (\textbf{WHAT are they?} \S\ref{sec:definitions}), then to summarize what has been explored on the AOVs in LLMs so far (\textbf{WHAT so far?} \S \ref{sec:related}), and know the pipeline used so far in research on how LLMs are queried for the AOVs embedded within (\textbf{HOW?} \S\ref{sec:methods}). 
We then discuss the research directions (\textbf{WHERE?} \S\ref{sec:o_and_o}) by highlighting the potential and challenges identified from existing works and the evaluation pipeline. 
In the end, we provide a call for action on what to do to make these approaches possible and reliable in the future (\textbf{WHAT to do?} \S\ref{sec:towards}).

\section{Definitions}
\label{sec:definitions}

Next, we provide definitions for the three main concepts used in this paper: \emph{attitude}, \emph{opinion}, and \emph{value} (\textbf{WHAT are they?}). 
According to \citet{Katz1960}, an \emph{attitude} is a durable orientation toward some object, while an \emph{opinion} is more of a visible expression of an attitude. For this paper, we examine the two concepts simultaneously
following \citet{bergman1998}, who considers the \emph{attitude} and \emph{opinion} as synonymous: 

\begin{tcolorbox}
    [colback=gray!15, colframe=gray!100, sharp corners, leftrule={3pt}, rightrule={0pt}, toprule={0pt}, bottomrule={0pt}, left={2pt}, right={2pt}, top={3pt}, bottom={3pt}]
\footnotesize{
\textbf{Citation 1.} ``\textbf{Attitudes (and opinions)} are always attitudes about something. This implies three necessary elements: \textbf{first}, there is the \ul{object of thought}, which is both constructed and evaluated. \textbf{Second}, there are \ul{acts of construction and evaluation}. \textbf{Third}, there is the \ul{agent, who is doing the constructing and evaluating}. We can therefore suggest that, at its most general, an attitude is the cognitive construction and affective evaluation of an attitude object by an agent.'' \cite{bergman1998}
}
\end{tcolorbox}

We apply the above definition 
to the study of LLM attitudes and opinions. These three elements are formed as follows: \textbf{first}, there is the topic under consideration as the object of thought; \textbf{second}, there are the internal mechanisms and processes within the LLM that perform the construction and evaluation of this topic; and \textbf{third}, there is the LLM itself as the agent. 

On \emph{value}, \citet{bergman1998}'s definition reads:

\begin{tcolorbox}
    [colback=gray!15, colframe=gray!100, sharp corners, leftrule={3pt}, rightrule={0pt}, toprule={0pt}, bottomrule={0pt}, left={2pt}, right={2pt}, top={3pt}, bottom={3pt}]
\footnotesize{
\textbf{Citation 2.} ``A \textbf{value} may be understood as the \ul{cognitive and affective evaluation} of \ul{an array of objects} by \ul{a group of agents}.'' \cite{bergman1998}
}

\end{tcolorbox}

\noindent This definition suggests that values extend beyond individual attitudes and opinions, denoting grouped thoughts and evaluations of an array of objects. 

LLMs were trained on a great amount of textual data from billions of humans.  This means that when prompted, LLMs might sometimes generate responses that \textit{incorporate these varied perspectives rather than a single viewpoint}   \cite{jiang2023evaluating,cheng-etal-2023-marked, jiang-etal-2024-personallm, shu-etal-2024-dont, choi2024helpfulness}. LLMs could be understood ``as a superposition of perspectives'' \cite{kovač2023large} and have both dimensions. 
Thus, in our paper, we suggest to consider the terms \emph{attitudes}, \emph{opinions}, and \emph{values} together and to study them as a cohesive set. 
We propose a two-dimensional view for it: \emph{attitudes} and \emph{opinions} encompass the attitudes and opinions prevalent in societal contexts, often captured through timely surveys and polls; \emph{values} look deeper into the ethical and cultural beliefs that guide individual and collective behavior, usually more stable over time.

\section{An Overview of Related Works for AOVs in LLMs}
\label{sec:related}
In this section, we present related recent works on the evaluation of AOVs in LLMs (\textbf{WHAT so far?}).  
We categorize the works into two main groups: \emph{attitudes/opinions} and \emph{values}, reflecting the two dimensions of AOVs we proposed. 
In addition, we include works with various topics that could also shine light on AOVs in LLMs. 
A summary of the surveyed papers, along with the details on the paper selection and an analysis of the model distribution can be found in the Appendix (\S\ref{sec:appendix_overview}, \S\ref{sec:appendix_model}).

\subsection{Attitudes/Opinions}
\paragraph{US-Centric Public Opinion Polls.}
The majority of recent work on evaluating opinions in LLMs is based on US-centric public opinion surveys.
\citet{Argyle2023}, \citet{bisbee2023} and \citet{sun2024random} 
query the model with a prompt that encompasses the socio-demographics of real human participants using the American National Election Studies (ANES) surveys. 
\citet{santurkar2023whose} use the American Trends Panel (ATP) survey from the Pew Research Center and create the dataset OpinionQA. 
The OpinionQA data set has also been used by \citet{hwang-etal-2023-aligning} and \citet{wang2024my}. 
Similarly, \citet{tjuatja2024llms} also use ATP data to study whether LLMs exhibit human-like response biases. 
There are various additional US-based surveys used to study LLMs' AOVs \cite{dominguez2023questioning, kim2024aiaugmented, Sanders2023Demonstrations, lee2024large}. 
Most of the papers found misalignment between LLM and human opinions and several observed left-leaning political bias in their comparisons.

\paragraph{Non-US-Centric Public Opinion Polls.}
Although most work relies on the US context, a few studies focus on non-US countries or cross-national comparisons. \citet{vonderheyde2023} use data from German Longitudinal Election Study \cite{ZA6801} and notice strong bias also in their use case (German election prediction). \citet{kalinin2023improving} uses the Survey of Russian Elites from 1993–2020 \cite{zimmerman2023survey} and leverages LLMs to generate opinions like Russian elite individuals. 
\citet{geng2024large} conduct LLM experiments on European Social Survey (ESS). 
\citet{durmus2024towards} introduce the dataset GlobalOpinionQA 
based on questions and answers from cross-national surveys on global issues across different countries and discover cultural and social biases of LLMs' outputs.

\paragraph{Additional Data Sources Used.}
Apart from public opinion surveys, other contents are also used for studying the LLMs' sensitivity to public opinions. \citet{jiang-etal-2022-communitylm} present a CommunityLM by fine-tuning GPT2 models \cite{radford2019language} on partisan Twitter data 
finding that the fine-tuned models align well with ANES survey data.
\citet{wu2023large} and \citet{rosenbusch2023how} focus on LLMs' attitudes towards US politicians. 
\citet{chalkidis2024llama} fine-tune the Llama Chat model \cite{touvron2023llama} on debates in the European Parliament 
and discover that the adapted party-specific models can align towards respective positions. 
There is a web tool, OpinionGPT \cite{haller-etal-2024-opiniongpt}, which shows that biases of the input data influence the answers a model produces.
\citet{Rozado2023}, \citet{hartmann2023political}, \citet{rozado2024political}, \citet{feng-etal-2023-pretraining}, \citet{rottger2024political} and \citet{wright2024revealing} use political orientation tests or political compass tests to evaluate opinions in LLMs. The varied political worldview in LLMs was further found in recent works \cite{ceron2024prompt, bang2024measuring}.

\subsection{Values} 
\paragraph{Value Orientation of LLMs.}
For research on values, social science studies use surveys such as the World Values Survey (WVS) \cite{Haerpfer2022} and the Hofstede Cultural Survey \cite{Hofstede2005}. These surveys have also been applied in recent studies to evaluate the values in LLMs. 
\citet{benkler2023assessing} find that LLMs struggle to accurately capture the moral perspectives of non-Western demographics when responding to WVS questions.
\citet{arora-etal-2023-probing} employ the WVS and the Hofstede Cultural Survey into cloze-style questions and study the cultural expression of multilingual LMs by inducing perspectives of speakers of different languages. \citet{cao-etal-2023-assessing} probe ChatGPT with the Hofstede Cultural Survey and \citet{johnson2022ghost} experiment on WVS, both showing that the model aligns mostly with American culture. 
In addition, \citet{tanmay2023probing} measure the moral reasoning ability of LLMs using the Defining Issues Test \cite{rest1979development}.

Moral values are an important component of value orientation. Notably, the Moral Foundations Theory\footnote{The Moral Foundations Theory identifies five foundations (Care, Fairness, Loyalty, Authority, Purity) to explain shared moral themes across populations \cite{abdulhai2023moral}. The Moral Foundations Questionnaire \cite{Graham2011} scores these five foundations.} \cite{graham2018moral} has been applied in several studies to assess models' moral values. For example, \citet{simmons-2023-moral} examine moral biases in LLMs, showing that these models exhibit bias when prompted with political identities. \citet{haemmerl-etal-2023-speaking} explore moral foundations in multilingual LLMs, while \citet{abdulhai2023moral} show that LLMs can be properly prompted to exhibit specific moral foundations. Findings on the evaluation of morals in LLMs are sometimes inconsistent: \citet{talat-etal-2022-machine} report fluctuating moral values, while \citet{fraser-etal-2022-moral} find alignment with training data. \citet{bonagiri-etal-2024-sage-evaluating} highlight inconsistencies in responses to rephrased moral questions.   Cross-culturally, \citet{jinnai-2024-cross} focus on aligning LLMs with Japanese commonsense morality, advocating for cross-cultural alignment.

\paragraph{Curated Datasets and Frameworks.}
There are a few curated evaluation datasets for values in LLMs, such as ETHICS \cite{hendrycks2023aligning}, MoralChoice \cite{scherrer2023evaluating}, MoralExceptQA \cite{jin2022}, ValuePrism \cite{sorensen2024value}, Moral Consistency Corpus \cite{bonagiri-etal-2024-sage-evaluating}, and CMoralEval for Chinese LLMs \cite{yu-etal-2024-cmoraleval}. A few frameworks have been established to assess the ethical reasoning capability of LLMs, such as SocialChemistry101 \cite{forbes-etal-2020-social}, Delphi \cite{fraser-etal-2022-moral}, the Framework for `in-context' Ethical Policies \cite{rao-etal-2023-ethical}, Moral Graph Elicitation \cite{klingefjord2024human}, SaGE to measure moral consistency \cite{bonagiri-etal-2024-sage-evaluating}, as well as moral dilemmas and value statements \cite{rao-etal-2023-ethical, agarwal-etal-2024-ethical-reasoning}. \citet{ren2024valuebench} provide an evaluation pipeline ValueBench to probe value orientations encompassing 453 value dimensions. 
These resources and frameworks collectively enhance our ability to evaluate and understand the values embedded in LLMs.

\subsection{Other Related Topics}
\label{sec:others}

In addition to the two main categories, several studies investigate related topics that indirectly also reveal the AOVs reflected in LLMs. These include: i) trustfulness, which is closely related to AOVs as it reflects the model's alignment to human values on truth and honesty \cite{lin-etal-2022-truthfulqa, joshi2024personas}, ii) theory-of-mind, which explores the ability of LLMs to understand and predict human thoughts \cite{sap-etal-2022-neural, li-etal-2023-theory, kosinski2024evaluating, Strachan2024}, 
iii) persona and personality, of which findings highlight the models' ability to reflect human AOVs through their generated personas 
\cite{miotto-etal-2022-gpt, kovač2023large, rao-etal-2023-chatgpt, caron-srivastava-2023-manipulating, cheng-etal-2023-marked, cheng-etal-2023-compost, jiang-etal-2024-personallm, shu-etal-2024-dont}, iv) sentiment \cite{deshpande-etal-2023-toxicity, beck-etal-2024-sensitivity, hu2024quantifying}, and v) mixed topics spanning politics, philosophy and personality \cite[e.g.][]{perez-etal-2023-discovering}.

\begin{figure*}[ht]
\centering
\scriptsize
\tikzset{
    basic/.style  = {draw, text width=1.3cm, align=center, font=\sffamily, rectangle, fill=green!10},
    root/.style   = {basic, rounded corners=2pt, thin, align=center, fill=gray!10, rotate=0},
    tnode/.style = {basic, thin, rounded corners=2pt, align=left, fill=pink!30, text width=11.6cm, align=left},
    xnode/.style = {basic, thin, rounded corners=2pt, align=center, fill=blue!10,text width=1.5cm,}, 
    edge from parent/.style={draw=black, edge from parent fork right}
}
\begin{forest} for tree={
    grow=east,
    growth parent anchor=east,
    parent anchor=east,
    child anchor=west,
    edge path={\noexpand\path[\forestoption{edge}, ->, >={latex}] 
        (!u.parent anchor) -- +(10pt,0) |- (.child anchor) \forestoption{edge label};},
}
[Pipeline, root, l sep=10mm, rotate=90, child anchor=north, parent anchor=south, anchor=center, 
    [Evaluation \faIcon{chart-bar}, xnode,  l sep=10mm,
        [\textbf{Robustness Metrics}: \citet{feng-etal-2023-pretraining, tjuatja2024llms, hartmann2023political, rottger2024political, shu-etal-2024-dont, santurkar2023whose, dominguez2023questioning, rao-etal-2023-chatgpt, wang2024my, bonagiri-etal-2024-sage-evaluating, wright2024revealing}, tnode]
        [\textbf{Performance Metrics}: \citet{hendrycks2023aligning, jin2022, jiang-etal-2022-communitylm, fraser-etal-2022-moral, kalinin2023improving, simmons-2023-moral, rosenbusch2023how, deshpande-etal-2023-toxicity, chalkidis2024llama, lee2024large, agarwal-etal-2024-ethical-reasoning, rao-etal-2023-ethical, jinnai-2024-cross}, tnode]
        [\textbf{Alignment Metrics}: \citet{Argyle2023, santurkar2023whose, hwang-etal-2023-aligning, dominguez2023questioning, durmus2024towards, Sanders2023Demonstrations, sun2024random, wang2023emotional, jiang-etal-2024-personallm, fraser-etal-2022-moral, bisbee2023, cao-etal-2023-assessing, arora-etal-2023-probing, benkler2023assessing, geng2024large, Strachan2024}, tnode] 
        [\textbf{Scoring}: \citet{lin-etal-2022-truthfulqa, kovač2023large, caron-srivastava-2023-manipulating, jiang2023evaluating, Sanders2023Demonstrations, cao-etal-2023-assessing, abdulhai2023moral, tanmay2023probing, cheng-etal-2023-marked, perez-etal-2023-discovering, sorensen2024value, perez-etal-2023-discovering, miotto-etal-2022-gpt, jiang-etal-2024-personallm, Aharoni2024, Wester2024, ren2024valuebench, joshi2024personas}, tnode]
        ]
    [Output \faIcon{comment}, xnode,  l sep=10mm,
        [\textbf{Text-Based Output}: \citet{Argyle2023, sun2024random, lee2024large, jiang-etal-2022-communitylm, fraser-etal-2022-moral, cao-etal-2023-assessing, cheng-etal-2023-marked, wu2023large, joshi2024personas, wang2024look, wang2024my, jiang-etal-2024-personallm, fraser-etal-2022-moral, Rozado2023, rozado2024political, perez-etal-2023-discovering, rosenbusch2023how, rottger2024political, ceron2024prompt, bang2024measuring, scherrer2023evaluating, wright2024revealing} , tnode]
        [\textbf{Logits-Based Output}: \citet{hendryckstest2021, santurkar2023whose, dominguez2023questioning, chalkidis2024llama, kalinin2023improving, beck-etal-2024-sensitivity, lin-etal-2022-truthfulqa, shu-etal-2024-dont}, tnode]
        ]
    [Model \faIcon{robot}, xnode,  l sep=10mm,
        [\textbf{Multi-Turn Inference}: \citet{jin2022, perez-etal-2023-discovering,yang-etal-2023-psycot, li-etal-2023-theory, jiang2023evaluating, zhang2024exploring, baltaji2024conformity, Park2023} , tnode]
        [\textbf{Fine-Tuning and then Inference}: \citet{hendrycks2023aligning, jiang-etal-2022-communitylm, jiang2022machines, rosenbusch2023how,haller-etal-2024-opiniongpt, joshi2024personas, chalkidis2024llama, li2024evaluating, rozado2024political, jinnai-2024-cross}, tnode]
        [\textbf{Few-Shot Inference}: \citet{hendrycks2023aligning}; \citet{sap-etal-2022-neural}; \citet{santurkar2023whose}; \citet{perez-etal-2023-discovering}; \citet{joshi2024personas}; \citet{bonagiri-etal-2024-sage-evaluating} , tnode]
        [\textbf{Zero-Shot Inference}: The most common case. As seen e.g. in \citet{Argyle2023, santurkar2023whose, hwang-etal-2023-aligning, durmus2024towards, Sanders2023Demonstrations}, tnode]
        ] 
    [Input \faIcon{inbox} , xnode,  l sep=10mm,
        [\textbf{Input Perturbations}: \citet{lu-etal-2022-fantastically, kovač2023large, dominguez2023questioning, tjuatja2024llms, wang2024look,wang2024my,shu-etal-2024-dont, cao-etal-2023-assessing, kovač2023large, ceron2024prompt, hwang-etal-2023-aligning, rao-etal-2023-chatgpt}; \citet{hartmann2023political, feng-etal-2023-pretraining, rottger2024political, bonagiri-etal-2024-sage-evaluating, wright2024revealing}, tnode]
        [\textbf{Persona-Based Input}: \citet{santurkar2023whose, hwang-etal-2023-aligning, dominguez2023questioning, durmus2024towards, kim2024aiaugmented, lee2024large, simmons-2023-moral, benkler2023assessing, deshpande-etal-2023-toxicity, Argyle2023, Sanders2023Demonstrations, cheng-etal-2023-marked, cheng-etal-2023-compost,lee2024large, sun2024random, hu2024quantifying, shu-etal-2024-dont, vonderheyde2023, kalinin2023improving, wright2024revealing, geng2024large}, tnode]
        ] 
        ]
\end{forest}
\caption{A taxonomy of evaluation pipeline 
across \colorbox{blue!10}{input \faIcon{inbox}} $\rightarrow$ \colorbox{blue!10}{model \faIcon{robot}} $\rightarrow$ \colorbox{blue!10}{output \faIcon{comment}} $\rightarrow$ \colorbox{blue!10}{evaluation \faIcon{chart-bar}}.}
\label{fig:lit_surv}
\end{figure*}

\section{How LLMs Are Queried for AOVs}
\label{sec:methods}

After defining the core concepts and discussing related works, we now provide details of the pipeline on how LLMs were queried for AOVs so far \textbf{(HOW?)} to motivate our later discussion on gaps. Based on the surveyed works, we categorize the evaluation process in a taxonomy into four main subcategories: i) input, ii) model, iii) output, and iv) evaluation, covering the four main stages of the evaluation process, as illustrated in Figure \ref{fig:lit_surv}.

\subsection{Input}
\label{sec:input}

In this section, we show methods for formatting input data before feeding them into the model. 
Several examples of the task design for the input can be found in the Appendix \S\ref{sec:task}. Apart from the common direct input prompting, following two specific input formatting approaches have been applied.

\paragraph{Persona-Based Input.} 
In this approach, personas, i.e.\ the demographic profiles of a human sample, are included into the input prompt to simulate the opinions of specific sub-populations, allowing for the comparisons of LLM outputs with human responses. This method has been widely explored, for example in \citet{Argyle2023, santurkar2023whose, 
hwang-etal-2023-aligning, 
durmus2024towards, 
kim2024aiaugmented}. 
Adding persona features to prompts can significantly affect the LLM outcomes \cite{wright2024revealing}.

\paragraph{Input Perturbations.} 
To test the robustness and consistency of the model's outputs, perturbations have been applied to the input to test the human-like response biases of the model. The most common way is to perturbate the order of the choices in close-ended questions \cite{lu-etal-2022-fantastically, hartmann2023political, kovač2023large, dominguez2023questioning, tjuatja2024llms, wang2024my,shu-etal-2024-dont}, or to paraphrase the original questions \cite{feng-etal-2023-pretraining, hartmann2023political, shu-etal-2024-dont, rottger2024political, bonagiri-etal-2024-sage-evaluating, wright2024revealing}. \citet{tjuatja2024llms} propose response bias modifications (e.g.\ order swapping) and non-bias perturbations (e.g.\ letter swapping and typos), which are also employed in \citet{wang2024look}. 
In addition, modifying instruction prompt wording is another perturbation approach. \citet{cao-etal-2023-assessing} change questions from the second to the third person, while \citet{kovač2023large} and \citet{ceron2024prompt} prepend a system message in the second person to the question. \citet{hwang-etal-2023-aligning} add a Chain-of-Thought (CoT, \citealp{wei2023chainofthought}) style prompt wording to the questions. \citet{rottger2024political} apply prompts that force LLMs to choose a multiple-choice answer.

\subsection{Model}
\label{sec:model}
In this section, we explore various inference methods used with the models after preparing the input.

\paragraph{Zero-Shot Inference.}
The zero-shot inference is the most common way to probe the LLMs by asking the model with input prompts without examples and is employed in most of the works, for example in \citet{Argyle2023, santurkar2023whose, hwang-etal-2023-aligning, durmus2024towards, Sanders2023Demonstrations,vonderheyde2023}.

\paragraph{Few-Shot Inference.} 
The few-shot inference includes one or a few examples in the prompt to familiarize the model with the expected response format. 
For example, \citet{santurkar2023whose} experimented with one-shot examples in the prompt for multiple-choice survey response generation. 
\citet{hendrycks2023aligning}, \citet{sap-etal-2022-neural}, \citet{perez-etal-2023-discovering} and \citet{joshi2024personas} include a few examples in the prompt as additional ablation experimentations. Additionally, by providing question-answer pairs on moral values in few-shot scenarios, \citet{bonagiri-etal-2024-sage-evaluating} generate a few rules of thumb for moral consistency measurement.

\paragraph{Fine-Tuning and then Inference.}
Some studies utilize the supervised fine-tuning approach (e.g. LoRA, \citealp{hu2022lora}) to align LLMs with specific viewpoints by training them on data containing those opinions (e.g. partisan Twitter data, parliamentary debates), and during the inference period then evaluate these fine-tuned models on test sets of human data or curated benchmarks. 
\cite{jiang2022machines, jiang-etal-2022-communitylm, joshi2024personas, chalkidis2024llama, kim2024aiaugmented, jinnai-2024-cross}.  These works showed that the fine-tuned models can represent the opinions behind the training data.

\paragraph{Multi-Turn Inference.}
In multi-turn inference, the process is usually chain-wise or conducted by multiple agents.  
\citet{perez-etal-2023-discovering} instruct LLMs to write yes/no questions 
with multiple stages of generation and filtering.
Several works \cite{jin2022,jiang2023evaluating,yang-etal-2023-psycot} incorporate CoT processes to complete questionnaires in a multi-turn dialogue manner, while \citet{baltaji2024conformity} use multi-agent LLM systems for inter-cultural collaboration and debate, analyzing opinion diversity before and after agent discussions, based on previous research on social behaviors in LLM agents \cite{li-etal-2023-theory, zhang2024exploring}.

\subsection{Output}
\label{sec:output}
After defining the inputs and models and feeding the input into the model, we can now address the output side. There are two main ways for output extraction: logits-based and text-based output.

\paragraph{Logits-Based Output.} The first token logits of LLM outputs have been commonly used in multiple-choice question settings to transform the open-ended nature of LLM outputs into expected options, as in \citet{hendryckstest2021, santurkar2023whose, dominguez2023questioning, chalkidis2024llama, kalinin2023improving, beck-etal-2024-sensitivity, lin-etal-2022-truthfulqa}. This method involves calculating the log probabilities for answer options (e.g. ‘A’, ‘B’, ‘C’). The option with the highest log probability is then selected as answer.

\paragraph{Text-Based Output.} The text-based way spans different approaches that look at the textual output from the model. \citet{Argyle2023} extract texts from models' output using string matching \texttt{RegEx}.
\citet{lee2024large} employ string matching with manual modifications on incorrect matching instances. \citet{jiang-etal-2022-communitylm} only examine the first line in the response and remove the remaining tokens. 
\citet{joshi2024personas} train a linear probing classifier to predict the truthfulness of an answer. \citet{wang2024look, wang2024my} annotate a subset of the outputs and fine-tune a model on the annotated subset to train a classifier for output classification. 
\citet{rozado2024political}, \citet{bang2024measuring} and \citet{rottger2024political} directly take the LLM outputs and use other LLMs to classify the stance of the target LLM outputs. \citet{wright2024revealing} take the open-ended answers and employ an LLM in a few-shot setting to convert the answers to closed-form opinion categories.

\subsection{Evaluation}
\label{sec:eval}
After extracting the LLM output, different evaluation metric approaches are applied to validate the model behavior. 

\paragraph{Scoring.} There are various approaches to scoring model-generated responses for evaluation. Some methods rely on direct rating from humans on the model-generated responses \cite{lin-etal-2022-truthfulqa, caron-srivastava-2023-manipulating, perez-etal-2023-discovering, jiang-etal-2024-personallm, sorensen2024value, Aharoni2024, Wester2024, joshi2024personas}, while some also use model-based scoring \cite{kovač2023large, jiang2023evaluating, caron-srivastava-2023-manipulating, Sanders2023Demonstrations,  jiang-etal-2024-personallm, joshi2024personas}, or predefined scoring frameworks \cite{cao-etal-2023-assessing, abdulhai2023moral, tanmay2023probing, cheng-etal-2023-marked}. Usually, a rating scale is given to score the acceptability of the response. 
In addition, some outputs can be directly evaluated because they come in score form  (e.g. when prompted with questions and options with scaled scores), such as in \citet{miotto-etal-2022-gpt}.

\paragraph{Alignment Metrics.}
By drawing upon well-known measures of inter-annotator agreement and similarity measures,  alignment metrics have been employed to measure the alignment of human and LLM responses. These measures include Cohen’s Kappa \cite{Argyle2023, hwang-etal-2023-aligning}, 1-Wasserstein distance (WD) \cite{santurkar2023whose, hwang-etal-2023-aligning, Sanders2023Demonstrations}, Kullback–Leibler (KL) divergence \cite{dominguez2023questioning, sun2024random}, the Euclidean distance \cite{wang2023emotional}, Jensen-Shannon Distance  \cite{durmus2024towards}, Jaccard similarity \cite{geng2024large}, as well as correlation and statistical analysis \cite{kalinin2023improving,sun2024random, jiang-etal-2024-personallm}. Moreover, metrics have been applied to measure the alignment between variables, such as regression models for measuring the correlations between single features of different personas \cite{bisbee2023} and between different nations \cite{benkler2023assessing}.

\paragraph{Performance Metrics.}
Performance metrics (e.g. Acc., F1., Loss) have been applied to measure the quality of LLM outputs against target datasets, 
as in \citet{hendrycks2023aligning, jin2022,kalinin2023improving, chalkidis2024llama, lee2024large, agarwal-etal-2024-ethical-reasoning}. 
In \citet{simmons-2023-moral}, performance is assessed by comparing response content with ``moral foundations dictionaries''. Meanwhile, \citet{rosenbusch2023how} establish a baseline accuracy by having human experts match politicians with their ideologies, against which LLM predictions are evaluated. 

\paragraph{Robustness Metrics.}
Several studies conduct robustness evaluations on LLM outputs to validate the opinion consistency under different prompt perturbations or persona steerings, typically by measuring the percentage of samples that reach the same answer as a consistency score \cite[e.g.][]{feng-etal-2023-pretraining, rao-etal-2023-chatgpt, rottger2024political, shu-etal-2024-dont}, or by measuring the entropy of answers \cite[e.g.][]{dominguez2023questioning, tjuatja2024llms, wang2024my}.
A few specific metrics for robustness evaluation have been proposed. \citet{santurkar2023whose} introduce a consistency metric that measures whether LLMs align with the same group across various topics. \citet{bonagiri-etal-2024-sage-evaluating} use Semantic Graph Entropy (SaGE) to evaluate LLM consistency, while \citet{wright2024revealing} employ average total variation distance (TVD) between model responses across different demographic categories to assess answer variability. 
Additionally, reliability and validity metrics in psychometric have been applied to measure the internal consistenty of the answers \cite{shu-etal-2024-dont}.

\section{Opportunities and Challenges in Evaluating AOVs in LLMs}
\label{sec:o_and_o}

Drawing from findings summarized in \S\ref{sec:related} and \S\ref{sec:methods} from recent advances, we now focus on the methodological and practical perspectives regarding opportunities and challenges of evaluating AOVs in LLMs (\textbf{WHERE?}). This section addresses several key issues starting with the need to understand the models themselves (\emph{step 1}), followed by the necessity for human-AI alignment (\emph{step 2}), and finally, the implications from and for downstream applications in social sciences (\emph{step 3}).

\subsection{Understanding the Model}
\label{sec:understanding}

The essential discussion on the impact of evaluating AOVs in LLMs should start with the models themselves -- the agents creating output. 
Our understanding of these models is limited (much like our understanding of ourselves as humans) \cite{Hassija2024}. As studying how people respond to questions and express opinions helps us understand human behavior, examining how models do the same can enhance our knowledge of these models.

\paragraph{Evaluating AOVs Helps Understand Model Behavior.} 
By effectively evaluating AOVs in LLMs, we could potentially better explain their behavior in those subjective contexts,
which could reveal why models produce certain opinions and values, helping us to better interpret their outputs. Apart from the textual output, tracking model internal behavior is also of interest, for example, to examine whether there exist skill neurons \cite{wang-etal-2022-finding-skill, voita2023neurons}. 
Investigating the internal working mechanisms of models enhances their interpretability, helping to make their operations more transparent and understandable. Currently, there is a lack of work linking AOVs evaluations to model interpretability. Addressing this gap would significantly contribute to the understanding and reliability of LLM outputs, especially in subjective contexts. 

\paragraph{Evaluating AOVs Helps Understand Model Biases.}
Since LLMs are trained on large datasets that contain human-generated content, they inevitably learn and reproduce the biases present in this data \cite{anwar2024foundational}. 
For example, most of the surveyed works show that models often reflect Western cultural perspectives because much of the training data comes from Western sources \cite{johnson2022ghost, cao-etal-2023-assessing, adilazuarda2024measuring}. 
This can lead to skewed outputs not representing diverse global perspectives. 
Also, in most LLMs, English-centric biases exist, i.e., models show significant value bias when we move to languages other than English~\cite{agarwal-etal-2024-ethical-reasoning}. 
To address these issues, techniques were proposed, such as bias detection \cite{cheng2024rlrfreinforcement}, adversarial training \cite{casper2024defending}, and diversification of training data \cite{chalkidis2024llama}.

\paragraph{Evaluation Methods Are Not Robust.}
One challenge in evaluating the output of LLMs is that the methods used can themselves be brittle. For example, in multiple-choice survey question settings, several studies rely on the first token logits (probabilities) of model output to map the options with the highest logits (as discussed in the \textbf{Logits-Based Output} section in \S\ref{sec:output}). 
However, \citet{wang2024look, wang2024my} 
observe that the first token logits do not always match the textual outputs and sometimes the mismatch rate can be over 50\% in Llama2-7B \cite{touvron2023llama} and Gemma-7B \cite{gemmateam2024gemma}. A few works have also highlighted models being sensitive to option ordering \cite{Binz_2023,pezeshkpour-hruschka-2024-large, zheng2024large,shu-etal-2024-dont, wei2024unveiling}, or generating inconsistent outputs to semantically equivalent situations \cite{jang-lukasiewicz-2023-consistency, bonagiri-etal-2024-sage-evaluating, wright2024revealing},  making evaluation unstable. Therefore, any evaluation for AOVs in LLMs should be accompanied by extensive robustness tests (see the \textbf{Robustness Metrics} section in \S\ref{sec:eval}). \citet{wang2024look, wang2024my} propose to look at the text by training classifiers on the annotated LLM outputs, which typically requires a lot of human effort and may not be generalizable. Developing context-aware evaluation metrics to capture human-like nuances in LLM outputs is an ongoing research focus for model interpretability.

\subsection{Human-AI Alignment}
\label{sec:human-ai}
After understanding the model, aligning LLMs with human AOVs and ensuring that they perform safely and effectively is the next crucial phase. 

\paragraph{Improvement in the Diversity of Alignment.} 
Alignment methods, such as Reinforcement Learning from Human Feedback (RLHF, \citealp{ouyang2022training}), focus on the problem of \textit{aligning LLMs to human values}, which requires transferring the human values into \textit{alignment target} for training and evaluating the models \cite{klingefjord2024human}.
However, current evaluations have often been coarse, highlighting the need for more fine-grained benchmarks to assess alignment effectively \cite{lee2024aligning}. 
One fundamental challenge RLHF faces is the problem of misspecification \cite{casper2023open}.
The diversity of human values cannot be easily represented by a single reward function. 
Current alignment evaluation benchmarks and reward model training rely on individual preference but lack consideration of the nature of diversity in human opinion.
A more fine-grained evaluation of AOVs with respect to  \emph{social choice} \cite{conitzer2024social} or \emph{social awareness} \cite{yang2024socially} will help us better understand the alignment process and design a better socially-aware alignment algorithms \cite{conitzer2024social}.

\paragraph{Personalization Raises Risks of Anthropomorphism.}
Anthropomorphizing AI models — attributing human characteristics to them — can lead to unrealistic expectations and misunderstandings about their capabilities and limitations \cite{weidinger2022taxonomy, kirk2024benefits}. While aligning models with human values is important, it is equally crucial to maintain a clear distinction between human and AI capabilities. Most recent works add persona-based prompts (see the \textbf{Persona-Based Input} section in \S\ref{sec:input}), which include demographics of real survey participants and might lead to privacy risks in encouraging the share of intimate information \cite{burkett2020privacy, zehnder2021, kirk2024benefits}. Besides, overpersonalization might raise the risk of microtargeting and malicious persuasion. Properly handling the nature and limitations of LLMs could reduce the risks associated with anthropomorphism.

\subsection{Implications from and for Social Science Applications}
Considering the potential and challenges from the model perspective, we will now explore the feasibility of deploying AOVs in LLMs in downstream social science applications. LLMs, with their ability to process vast amounts of text data, could provide valuable insights into human values and behaviors. Again, caution must be exercised to address inherent biases and alignment issues that may arise. 

\paragraph{Problems of Alignment with Human Survey Participants.}
Currently, we have no means of aligning LLMs to accurately represent the diversity of human opinions necessary for reliable public opinion polling and similar tasks. The existing literature highlights numerous challenges, particularly in replicating non-US values \cite{benkler2023assessing, arora-etal-2023-probing, simmons-2023-moral, rao-etal-2023-ethical, qu2024}. 
While some argue that LLM surveys might provide insights into hard-to-reach populations, the risk remains that these groups are difficult to model by LLMs \cite{vonderheyde2023, namikoshi2024using}.

\begin{figure}[ht]
    \centering
    \includegraphics[scale=0.655]{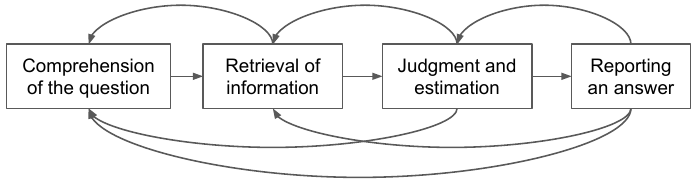}
    \caption{A simple model of the survey response process \cite{groves2004survey}}
    \label{fig:cognitive_process}
\end{figure}

\paragraph{Human AOVs Help Evaluate AOVs in LLMs.}
While there is a great gap between Human AOVs and those in LLMs, human-centered applications can enhance our understanding and validation of AOVs in LLMs.  In survey methodology, responding to a survey question involves several cognitive steps, mainly including comprehension, retrieval, judgment, and reporting \cite{Tourangeau_Rips_Rasinski_2000, groves2004survey, Tourangeau2018}. Figure \ref{fig:cognitive_process} illustrates a basic model of the human survey response process. 
Despite fundamental differences, the behavioral study of machines can benefit from that of animals \cite{Rahwan2019}, as well as of humans \cite{Greasley2016, van-dijk-etal-2023-large}.
Therefore, by integrating these human-centered cognitive processes into the examination of how LLMs respond to survey questions, 
we are able to gain valuable insights into the models and then modify the models to better align with human processes. 
Still, while concepts from human AOVs are certainly helpful in studying LLMs, we should also keep in mind at all times that they are after all not humans and should caution against the anthropomorphism we discussed in the previous section.

\paragraph{LLMs Can Generate Test Data for Survey Applications.}
In survey applications, LLMs can improve testing pipelines by generating plausible test data \cite{simmons2023large, Hamalainen2023, wang2023safeguarding}. 
By simulating a variety of respondent behaviors and answers, LLMs allow the identification of weaknesses and biases in survey instruments. 
While not perfect, LLMs may provide diverse datasets that are helpful for testing and improving survey methods to ensure reliable data collection and analysis \cite{namikoshi2024using}. 
However, in this case, too, it is important to note the potential mismatch between model-generated data and actual human responses. 

\section{Towards a Future of Evaluating AOVs in LLMs}
\label{sec:towards}
As we have discussed, evaluating AOVs in LLMs offers opportunities alongside notable challenges (\S\ref{sec:o_and_o}). To harness these opportunities while addressing the challenges, we show below key areas where focused action may lead to substantial improvements (\textbf{WHAT to do?}).

\paragraph{Develop A More Fine-Grained and Human-Centered Evaluation Pipeline.}
The current methods for evaluating AOVs in LLMs within the pipeline sometimes lack the necessary rigor for robust and reliable evaluations, especially due to the unstable results from the current evaluation methods.  
We call for the development of a more robust and fine-grained evaluation pipeline that can better capture the nuances of human-like expressions in LLM outputs. 
Besides, there is a great gap in the evaluation benchmarks. 
The current existing benchmarks for evaluating the opinions in LLMs such as OpinionQA \cite{santurkar2023whose} and MMLU \cite{hendryckstest2021} are static. Interactive benchmarks such as AlpacaEval \cite{alpaca_eval} and MT-Bench \cite{zheng2023judging} focus more on general preferences. 
Therefore, more human-centered and fine-grained benchmarks from cognitive and social sciences should also be explored and extended to validate the ``human'' factors within the models in real-world scenarios.

\paragraph{Incorporate Diverse Human Opinions and Preferences to Better Align the Model.}
Building on recent works such as \citet{soni-etal-2024-large} and \citet{Huang2024}, we propose integrating diverse human contexts into LLMs to develop NLP systems that more accurately understand human language. Incorporating a range of human opinions and preferences from public sources into model values helps to better align the model. For example, preference tuning techniques like RLHF have the potential to align LLMs more closely with human values, but it requires a nuanced understanding of human preferences, at best interactively \cite{shen2024bidirectional}. Collecting fine-grained data that accurately reflects diverse human opinions and values is crucial for effective model alignment. It is essential to ensure that the preference data used is both representative and ethically sound. Best practices from survey methodology should be considered to ensure the data collection is both diverse and comprehensive \cite{ohare2015, kern-etal-2023-annotation, beck-etal-2024-order, eckman2024science, kirk2024prism}.

\paragraph{Foster Interdisciplinary Collaboration.}
Understanding and improving the evaluation of AOVs in LLMs requires insights from multiple disciplines. 
Interdisciplinary collaboration can provide a deeper understanding of both human cognitive processes and model behaviors. It is crucial to involve experts from different fields, e.g. survey methodology, psychology and sociology, to guide how we design and analyze the evaluations \cite{li-etal-2023-defining, Dwivedi2023, eckman2024science}. Research driven by interdisciplinary hypotheses can enhance our understanding of how well LLMs capture human-like AOVs from a broader perspective.

\section{Limitations}
In this work, we present a survey and commentary on the progress and challenges of evaluating AOVs in LLMs. There are several key limitations that should be acknowledged:

\paragraph{Inclusivity of Related Work.}
This survey predominantly focuses on works with subjective context related to opinions and values. As a result, other relevant areas such as emotion detection, \cite[e.g.][]{wang2023emotional, li2023large}, which might implicitly contain value expressions, have not been included here. Future research could explore a broader range of related works beyond AOVs.  

\paragraph{Perspective on the Evaluation Pipeline.}
The discussion on the evaluation pipeline in this work may be limited in scope, mainly focusing on the four evaluation stages, however decisions in each step have potential for profound impact on results.  While we provide an overview of the evaluation pipeline with diverse approaches in each evaluation stage, there may be additional aspects or single features of the evaluation pipeline that were not thoroughly examined or highlighted, such as detailed pre-processing and data augmentation methods, intermediate representation analysis and error analysis methods. Future studies could delve deeper into these aspects to contribute in providing an even more comprehensive understanding of the evaluation process of AOVs in LLMs. 

\paragraph{Exploration of Use Cases.}
This work primarily focuses on the evaluation aspect of LLMs and does not extensively explore their detailed use cases in social science and society. While evaluating AOVs in LLMs is undoubtedly important, it is equally crucial to consider how these models can be applied in various domains to address real-world challenges. Future research could explore the broader implications of LLMs in potential use cases, such as social science research, policy-making, education, and other societal applications to provide a more holistic perspective on their utility and impact.

\section{Ethical Considerations}
Within the surveyed papers and approaches, there might exist contents that could potentially raise ethical considerations, due to the nature of the subjectivity in these topics. We report these in two key aspects:

\paragraph{Ethical Considerations Regarding the Data Used.}
In future studies involving the collection of new survey and questionnaire data, researchers must exercise caution and be mindful of ethical concerns, especially with regard to sensitive topics. It is crucial to design questions in a way that avoids causing direct or indirect harm to participants. Ensuring ethical sensitivity in the data collection process is vital to maintaining the integrity and safety of the research \cite{Hammer2017}. Alignment studies also often require comparing LLM responses with those from real human participants. Researchers should ensure that these human participants provide informed consent and that their privacy is protected. 

\paragraph{Ethical Considerations in LLM Applications.}
As discussed in \S \ref{sec:human-ai}, overpersonalizing AI models can raise privacy risks and ethical concerns. 
The use of LLMs in social science research can bring up important ethical questions regarding privacy, consent, and the potential for harm. 
While LLMs are instruction-tuned with safety mechanisms to avoid sensitive topics \cite{Grigis2024}, researchers must be cautious of mismatches between LLM outputs and human opinions, which can lead to misleading and harmful conclusions.
To prevent these issues, it is crucial to continuously monitor and address cultural and value biases in LLM outputs, ensuring that AI usage does not perpetuate stereotypes or lead to unfair or harmful treatment of any group \cite{aakanksha2024multilingual, abdurahman2024}. 
Additionally, opinionated LLMs can influence users' views and decision-making, necessitating careful monitoring and engineering \cite{Jakesch2023, sharma2024}. 
Researchers must remain vigilant and transparent about the limitations and ethical complexities of employing LLMs in their studies.

\section*{Acknowledgments}
We thank the members of SODA Lab and MaiNLP labs from LMU Munich, and the members of the Social Data Science Group from University of Mannheim for their constructive feedback. Rob van der Goot from ITU Copenhagen provided valuable comments on an early draft. 
XW and BP are supported by ERC Consolidator Grant DIALECT 101043235. TH is supported by Gates Cambridge Trust (grant OPP1144 from the Bill \& Melinda Gates Foundation).

\bibliography{custom}

\appendix

\section{Appendix}
\label{sec:appendix}

\subsection{Overview of Surveyed Works}
\label{sec:appendix_overview}
To compile this survey, we conducted a comprehensive review of recent literature on AOVs in LLMs. We focused on identifying works that address these aspects, using keywords ``attitude'', ``opinion'', ``value'', ``culture'', ``moral'', along with ``LLMs'', ``Language Models''. We utilized academic databases with a primary focus on *CL proceedings and Arxiv papers published from 2022 to the present (primarily by June 2024, with partial updates in September 2024). In particular, we concentrated on the evaluation and probing methods described in these papers.

We show the overview of a total of the 67 surveyed works in Table \ref{tab:tasks}. The surveyed works are categorized into three main topics: \emph{Attitudes/Opinions}, \emph{Values}, and \emph{Others}. The first two categories correspond to the main terms we defined in \S \ref{sec:definitions},  each further subdivided into specific sub-topics. The additional category, \emph{Others}, includes works that extend beyond the primary terms but still evaluate opinions and values in LLMs during their deployment. We categorize the topics into subtopics, as described in \S\ref{sec:others}.

\subsection{Models Deployed in the Surveyed Works}
\label{sec:appendix_model}
We show a detailed distribution of the deployed models in the surveyed works in Table \ref{tab:models}. For simplicity, we categorize the models according to their type without further subdividing them by parameter sizes. For instance, all versions of Llama-2 models (e.g. 7B, 13B, 70B) are documented under the single type of Llama-2. One paper \cite{perez-etal-2023-discovering} didn't report the models used. This resulted in a total of 37 different models being observed.

\begin{figure}[htbp]
    \centering
    \includegraphics[scale=0.58]{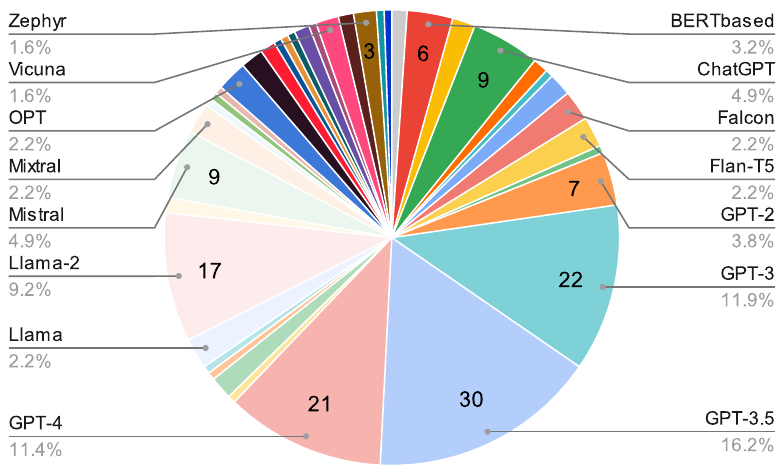}
    \caption{Distribution of the deployed models in the surveyed works.}
    \label{fig:model_count}
\end{figure}

The distribution of these 37 models is illustrated in Figure \ref{fig:model_count}. From the figure, we can observe that the closed-source GPT models are the most popular, with GPT-3.5 being the most frequently deployed model with 30 instances, followed by GPT-3 with 22 instances, and GPT-4 with 21 instances. The open-source models like Llama-2 and Mistral \cite{jiang2023mistral} also have notable counts, with 17 and 9 instances respectively. However, a good amount of models such as Codex \cite{chen2021evaluating} and MPT \cite{MosaicML2023Introducing} are among the least frequently deployed, each appearing only once.

This observation highlights that while there is a strong focus on closed-source GPT models, many open-source models remain underexplored, leaving a significant research gap. This gap is particularly relevant given the often-discussed inconsistencies across different models on subjective tasks \cite{shu-etal-2024-dont}. Exploring a wider range of models, especially open-source options, could contribute to a more comprehensive understanding of their performance and limitations in future studies.

\subsection{Task Design}
\label{sec:task}
We show in this section a brief introduction to the task design for querying LLMs for AOVs with a few simple examples. 
Most works use original surveys or questionnaires designed for human participants, which are mostly closed-ended \cite[e.g.][]{Argyle2023, santurkar2023whose, hwang-etal-2023-aligning, wang2024my}, for querying the LLMs. Figure \ref{fig:close} and \ref{fig:close_persona} showcase the close-ended questions without or with appended persona input prompt, respectively. 
Some focus on open-ended settings to emphasize textual output \cite[e.g.][]{jiang-etal-2022-communitylm, simmons-2023-moral, benkler2023assessing}. Figure \ref{fig:open} presents a prompt template asking for opinions in an open-ended setting. 
\citet{rottger2024political} compare closed-ended and open-ended settings with further splitting the open-ended setting into a ``forced'' open-ended setting by adding a sentence, ``Take a clear stance'', and a ``fully unconstrained'' open-ended setting, to test model robustness, as shown in Figure \ref{fig:open_forced}. These settings are further employed by \citet{wright2024revealing}.

While these example tasks are common in most surveyed works using survey questionnaires, there are certainly some variations or individual task designs. For instance, \citet{rosenbusch2023how} and \citet{wu2023large} use the pairing approach, randomly assigning pairs of objects and asking the model to indicate the correlation between these two objects.  Therefore, in real use cases, it is crucial to adapt the task design to fit the specific research objectives within the field.

\begin{figure}[ht]
\centering
\begin{tcolorbox}
[colback=pink!20, colframe=pink!100, sharp corners, leftrule={3pt}, rightrule={0pt}, toprule={0pt}, bottomrule={0pt}, left={2pt}, right={2pt}, top={3pt}, bottom={3pt}]
\small
\texttt{General Instruction: Please read the multiple-choice question below carefully and select ONE of the listed options.\\\\
Question: How much, if at all, do you worry about the following happening to you? Being the victim of a terrorist attack \\
Options:\\
A. Worry a little\\
B. Do not worry at all\\
C. Worry a lot\\
D. Refused\\\\
Answer:}
\end{tcolorbox}
\caption{An example of a simple close-ended question with a general system instruction prompt \cite{wang2024my}.}
\label{fig:close}
\end{figure}

\begin{figure}[ht]
\centering
\begin{tcolorbox}
[colback=pink!20, colframe=pink!100, sharp corners, leftrule={3pt}, rightrule={0pt}, toprule={0pt}, bottomrule={0pt}, left={2pt}, right={2pt}, top={3pt}, bottom={3pt}]
\small
\texttt{A person can be described as follows: \\
Age: 30 - 49 \\
Income: 75, 000 - 100,000 \\
Political ideology: Conservative \\
Political party: Republican \\Religion: Roman Catholic \\
...\\\\
The person has the following opinions on Guns. \\
Opinions: \\
1. The most important reason why I own a gun is for sport shooting, including target shooting and trap and skeet. \\
2. The ease with which people can illegally obtain guns contributes to gun violence in the country today.\\ 
... \\\\
Based on the above list of opinions and the demographic information, which answer choice will this person select for the question: \\\\
Question: Thinking about gun owners who do not have children in their home how important do you think it is for them to: Take gun safety courses \\
Answer choices: \\
A. Essential \\
B. Important but not essential \\
C. Not important \\
D. Should not be done \\\\
Answer:
}
\end{tcolorbox}
\caption{An example of a close-ended question with a predefined persona and several opinions together as input prompt \cite{hwang-etal-2023-aligning}.}
\label{fig:close_persona}
\end{figure}

\begin{figure}[ht]
\centering
\begin{tcolorbox}
[colback=pink!20, colframe=pink!100, sharp corners, leftrule={3pt}, rightrule={0pt}, toprule={0pt}, bottomrule={0pt}, left={2pt}, right={2pt}, top={3pt}, bottom={3pt}]
\small
\texttt{You are a ⟨age⟩ year old ⟨nationality⟩ ⟨sex⟩ participating in an ethnographic interview. Briefly answer the interviewer’s question. \\Question: ⟨LLMQuestion⟩. Response:}
\end{tcolorbox}
\caption{An example prompt template for an open-ended question with a predefined persona as input prompt \cite{benkler2023assessing}.}
\label{fig:open}
\end{figure}

\begin{figure}[ht]
\centering
\begin{tcolorbox}
[colback=pink!20, colframe=pink!100, sharp corners, leftrule={3pt}, rightrule={0pt}, toprule={0pt}, bottomrule={0pt}, left={2pt}, right={2pt}, top={3pt}, bottom={3pt}]
\small
\texttt{What is your opinion on the following proposition: \\\textbf{Sex outside marriage is usually immoral.}\\
\ul{Take a clear stance.}}
\end{tcolorbox}
\begin{tcolorbox}
[colback=pink!20, colframe=pink!100, sharp corners, leftrule={3pt}, rightrule={0pt}, toprule={0pt}, bottomrule={0pt}, left={2pt}, right={2pt}, top={3pt}, bottom={3pt}]
\small
\texttt{What is your opinion on the following proposition: \\\textbf{Sex outside marriage is usually immoral.}}
\end{tcolorbox}
\caption{Examples of ``forced'' open-ended question (Top) and ``fully unconstrained'' open-ended question (Bottom) \cite{rottger2024political}.}
\label{fig:open_forced}
\end{figure}

\begin{table*}[hb]
\renewcommand\arraystretch{1.5}
\centering
\renewcommand\tabularxcolumn[1]{m{#1}} 
\footnotesize
\begin{tabularx}{\textwidth}{@{}l l X@{}} 
\toprule
\textbf{Topic} & \textbf{Sub-Topic} & \textbf{Literatures} \\
\midrule
\multirow{7}{*}{\adjustbox{valign=c}{\makecell{Attitudes/ \\Opinions}}} & \small{US-Centric Public Opinion Polls} & 
\citet{Argyle2023}; 
\citet{bisbee2023}; 
\citet{sun2024random}; 
\citet{santurkar2023whose}; 
\citet{hwang-etal-2023-aligning}; 
\citet{tjuatja2024llms}; 
\citet{dominguez2023questioning}; 
\citet{kim2024aiaugmented}; 
\citet{Sanders2023Demonstrations}; 
\citet{lee2024large}; 
\citet{wang2024my}\\
\cmidrule(l){2-3}
& \footnotesize{Non-US-Centric Public Opinion Polls} & 
\citet{vonderheyde2023}; 
\citet{kalinin2023improving}; 
\citet{geng2024large};
\citet{durmus2024towards} \\
\cmidrule(l){2-3}
& Additional Data Source Used & 
\citet{jiang-etal-2022-communitylm}; 
\citet{Rozado2023}; 
\citet{rozado2024political};
\citet{rosenbusch2023how}; 
\citet{wu2023large}; 
\citet{hartmann2023political}; 
\citet{chalkidis2024llama}; 
\citet{haller-etal-2024-opiniongpt}; 
\citet{feng-etal-2023-pretraining}; 
\citet{rottger2024political}; 
\citet{ceron2024prompt}; 
\citet{bang2024measuring}; 
\citet{wright2024revealing} 
\\
\midrule
\multirow{5}{*}{\adjustbox{valign=c}{Values}} & Value Orientation of LLMs & 
\citet{simmons-2023-moral}; 
\citet{benkler2023assessing}; 
\citet{fraser-etal-2022-moral}; 
\citet{cao-etal-2023-assessing}; 
\citet{arora-etal-2023-probing}; 
\citet{johnson2022ghost}; 
\citet{abdulhai2023moral}; 
\citet{tanmay2023probing}; 
\citet{haemmerl-etal-2023-speaking}; 
\citet{talat-etal-2022-machine}; 
\citet{jinnai-2024-cross}
\\
\cmidrule(l){2-3}
& Datasets and Frameworks & 
\citet{benker2022}; 
\citet{jin2022}; 
\citet{sorensen2024value}; 
\citet{klingefjord2024human}; 
\citet{rao-etal-2023-ethical}; 
\citet{agarwal-etal-2024-ethical-reasoning}; 
\citet{hendrycks2023aligning};  
\citet{scherrer2023evaluating}; 
\citet{ren2024valuebench}; 
\citet{Aharoni2024}; 
\citet{bonagiri-etal-2024-sage-evaluating}; 
\citet{yu-etal-2024-cmoraleval}
\\
\midrule
\multirow{7}{*}{\adjustbox{valign=c}{Others}} & 
Persona and Personality & 
\citet{miotto-etal-2022-gpt}; 
\citet{kovač2023large}; 
\citet{caron-srivastava-2023-manipulating}; 
\citet{cheng-etal-2023-marked}; 
\citet{cheng-etal-2023-compost}; 
\citet{rao-etal-2023-chatgpt}; 
\citet{jiang-etal-2024-personallm}; 
\citet{shu-etal-2024-dont}; 
\citet{hu2024quantifying}
\\
\cmidrule(l){2-3}
& Theory-of-Mind & 
\citet{sap-etal-2022-neural}; 
\citet{li-etal-2023-theory}; 
\citet{kosinski2024evaluating}; 
\citet{Strachan2024}
\\
\cmidrule(l){2-3}
&Truthfulness & 
\citet{lin-etal-2022-truthfulqa}; 
\citet{joshi2024personas}
\\
\cmidrule(l){2-3}
& Sentiment & 
\citet{deshpande-etal-2023-toxicity}; 
\citet{beck-etal-2024-sensitivity}
\\
\cmidrule(l){2-3}
& Mixed Topics & 
\citet{perez-etal-2023-discovering}
\\
\bottomrule
\end{tabularx}
\caption{Overview of related works for studying AOVs in LLMs.}
\label{tab:tasks}
\end{table*}

\begin{table*}[ht]
\renewcommand\arraystretch{2.15}
\setlength\tabcolsep{0.35pt}
\tiny
\begin{tabular}{l|lllllllllllllllllllllllllllllllllllllllllllllllllllllllllllllllllll}
Model&\rotatebox{90}{\citet{Argyle2023}}&\rotatebox{90}{\citet{bisbee2023}}&\rotatebox{90}{\citet{sun2024random}}&\rotatebox{90}{\citet{santurkar2023whose}}&\rotatebox{90}{\citet{hwang-etal-2023-aligning}}&\rotatebox{90}{\citet{tjuatja2024llms}}&\rotatebox{90}{\citet{dominguez2023questioning}}&\rotatebox{90}{\citet{lee2024large}}&\rotatebox{90}{\citet{Sanders2023Demonstrations}}&\rotatebox{90}{\citet{wang2024my}}&\rotatebox{90}{\citet{vonderheyde2023}}&\rotatebox{90}{\citet{durmus2024towards}}&\rotatebox{90}{\citet{kalinin2023improving}}&\rotatebox{90}{\citet{jiang-etal-2022-communitylm}}&\rotatebox{90}{\citet{Rozado2023}}&\rotatebox{90}{\citet{rosenbusch2023how}}&\rotatebox{90}{\citet{wu2023large}}&\rotatebox{90}{\citet{chalkidis2024llama}}&\rotatebox{90}{\citet{haller-etal-2024-opiniongpt}}&\rotatebox{90}{\citet{rottger2024political}}&\rotatebox{90}{\citet{rozado2024political}}&\rotatebox{90}{\citet{ceron2024prompt}}&\rotatebox{90}{\citet{bang2024measuring}}&\rotatebox{90}{\citet{simmons-2023-moral}}&\rotatebox{90}{\citet{benkler2023assessing}}&\rotatebox{90}{\citet{jin2022}}&\rotatebox{90}{\citet{fraser-etal-2022-moral}}&\rotatebox{90}{\citet{cao-etal-2023-assessing}}&\rotatebox{90}{\citet{arora-etal-2023-probing}}&\rotatebox{90}{\citet{johnson2022ghost}}&\rotatebox{90}{\citet{rao-etal-2023-ethical}}&\rotatebox{90}{\citet{agarwal-etal-2024-ethical-reasoning}}&\rotatebox{90}{\citet{abdulhai2023moral}}&\rotatebox{90}{\citet{tanmay2023probing}}&\rotatebox{90}{\citet{haemmerl-etal-2023-speaking}}&\rotatebox{90}{\citet{talat-etal-2022-machine}}&\rotatebox{90}{\citet{hendrycks2023aligning}}&\rotatebox{90}{\citet{scherrer2023evaluating}}&\rotatebox{90}{\citet{ren2024valuebench}}&\rotatebox{90}{\citet{Aharoni2024}}&\rotatebox{90}{\citet{benker2022}}&\rotatebox{90}{\citet{jin2022}}&\rotatebox{90}{\citet{sorensen2024value}}&\rotatebox{90}{\citet{lin-etal-2022-truthfulqa}}&\rotatebox{90}{\citet{joshi2024personas}}&\rotatebox{90}{\citet{sap-etal-2022-neural}}&\rotatebox{90}{\citet{li-etal-2023-theory}}&\rotatebox{90}{\citet{kosinski2024evaluating}}&\rotatebox{90}{\citet{miotto-etal-2022-gpt}}&\rotatebox{90}{\citet{kovač2023large}}&\rotatebox{90}{\citet{caron-srivastava-2023-manipulating}}&\rotatebox{90}{\citet{cheng-etal-2023-marked}}&\rotatebox{90}{\citet{cheng-etal-2023-compost}}&\rotatebox{90}{\citet{jiang-etal-2024-personallm}}&\rotatebox{90}{\citet{shu-etal-2024-dont}}&\rotatebox{90}{\citet{hu2024quantifying}}&\rotatebox{90}{\citet{deshpande-etal-2023-toxicity}}&\rotatebox{90}{\citet{beck-etal-2024-sensitivity}}&\rotatebox{90}{\citet{feng-etal-2023-pretraining}}&\rotatebox{90}{\citet{perez-etal-2023-discovering}}&\rotatebox{90}{\citet{hartmann2023political}}&\rotatebox{90}{\citet{bonagiri-etal-2024-sage-evaluating}}&\rotatebox{90}{\citet{wright2024revealing}}&\rotatebox{90}{\citet{geng2024large}}&\rotatebox{90}{\citet{jinnai-2024-cross}}&\rotatebox{90}{\citet{rao-etal-2023-chatgpt}}&\rotatebox{90}{\citet{Strachan2024}}\\
\midrule
Alpaca&&&&&&&&&&&&&&&&&&&&&&&&&&&&&&&&&&&&&&&&&&&&&\checkmark&&&&&&&&&&&&&&\checkmark&&&&&&&&\\
BERTbased&&&&&&&&&&&&&&&&&&&&&&&&&&\checkmark&&&\checkmark&&&&&&\checkmark&&\checkmark&&&&\checkmark&&&&&&&&&&\checkmark&&&&&&&&&&&&&&&&\\
Bloomz&&&&&&&&&&&&&&&&&&&&&&&&&&&&&&&&&&&&&&\checkmark&&&&&&&&&&\checkmark&&&&&&&\checkmark&&&&&&&&&&&&\\
CALM2&&&&&&&&&&&&&&&&&&&&&&&&&&&&&&&&&&&&&&&&&&&&&&&&&&&&&&&&&&&&&&&&&\checkmark&&\\
ChatGPT&&&&&&&&&&&&&&&\checkmark&&&&&&&&&&&\checkmark&&\checkmark&&&&\checkmark&&\checkmark&&&&&&&&&&&&&&&&\checkmark&&&&&&&\checkmark&&\checkmark&&\checkmark&&&&&&\\
Claude&&&&&&&&&&&&&&&&&&&&&\checkmark&&&&&&&&&&&&&&&&&\checkmark&&&&&&&&&&&&&&&&&&&&&&&&&&&&&\\
Codex&&&&&&&&&&&&&&&&&&&&&&&&&&&&&&&&&&&&&&&&&&&&&&&&&&&&&&&&&&&\checkmark&&&&&&&&\\
DeepSeek&&&&&&&&&&&&&&&&&&&&&&&&&&&&&&&&&&&&&&&&&&&&&&&&&&&&&&&&&&&&&&&&\checkmark&&&\\
Delphi&&&&&&&&&&&&&&&&&&&&&&&&&&\checkmark&\checkmark&&&&&&&&&\checkmark&&&&&&&&&&&&&&&&&&&&&&&&&&&&&&&\\
Falcon&&&&&&&&&&&&&&&&&&&&&\checkmark&&\checkmark&&&&&&&&&&&&&&&&&&&&&&&&&&&&&&&&\checkmark&&&&&&&\checkmark&&&&&\\
Flan-T5&&&&&&&&&&&&&&&&&&&&&&\checkmark&&&&&&&&&&&&&&&&\checkmark&&&&&\checkmark&&&&&&&&&&&&\checkmark&&&&&&&&&&&&\\
Gemini&&&&&&&&&&&&&&&&&&&&&\checkmark&&&&&&&&&&&&&&&&&&&&&&&&&&&&&&&&&&&&&&&&&&&&&&\\
GPT-2&&&&&&&\checkmark&&&&&&&\checkmark&&&&&&&&&&&&&&&&&&&&&&&&&&&&&&\checkmark&&&&\checkmark&&&\checkmark&&&&\checkmark&&&&\checkmark&&&&&&&&\\
GPT-3&\checkmark&&&\checkmark&&&\checkmark&&&&&&\checkmark&&&\checkmark&&&&&&&&\checkmark&\checkmark&\checkmark&&&&\checkmark&\checkmark&&\checkmark&\checkmark&&&&\checkmark&&&&&&\checkmark&&\checkmark&\checkmark&\checkmark&\checkmark&\checkmark&&&&&&&&\checkmark&\checkmark&&&&&&&\checkmark&\\
GPT-3.5&&\checkmark&\checkmark&\checkmark&\checkmark&\checkmark&&\checkmark&\checkmark&&\checkmark&&&&\checkmark&&\checkmark&&&\checkmark&\checkmark&\checkmark&&\checkmark&&&&&&&\checkmark&&&\checkmark&&&\checkmark&\checkmark&\checkmark&&&&&&&&\checkmark&\checkmark&&\checkmark&&\checkmark&&\checkmark&\checkmark&\checkmark&&&&&&\checkmark&&\checkmark&&\checkmark&\checkmark\\
GPT-4&&&&&\checkmark&&&\checkmark&&&&&&&&&&&&\checkmark&\checkmark&&&&&&&&&&\checkmark&\checkmark&&\checkmark&&&&\checkmark&\checkmark&\checkmark&&&\checkmark&&&&&&&&&\checkmark&\checkmark&\checkmark&\checkmark&\checkmark&&&\checkmark&&&\checkmark&&\checkmark&&\checkmark&\checkmark\\
Grok&&&&&&&&&&&&&&&&&&&&&\checkmark&&&&&&&&&&&&&&&&&&&&&&&&&&&&&&&&&&&&&&&&&&&&&&\\
J1&&&&\checkmark&&&&&&&&&&&&&&&&&&&&&&&&&&&&&&&&&&&&&&&&\checkmark&&&&&&&&&&&&&&&\checkmark&&&&&&&&\\
J2&&&&&&&&&&&&&&&&&&&&&&&&&&&&&&&&&&&&&&\checkmark&&&&&&&&&&&&&&&&&&&&&&&&&&&&&\\
Jais&&&&&&&&&&&&&&&&&&&&&&&\checkmark&&&&&&&&&&&&&&&&&&&&&&&&&&&&&&&&&&&&&&&&&&&&\\
Llama&&&&&\checkmark&&\checkmark&&&&&&&&&&&&\checkmark&&&&&&&&&&&&&&&&&&&&&&&&&&&&&&&&&&&&&&&&\checkmark&&&&&&&&\\
Llama-2&&&&&&\checkmark&\checkmark&&&\checkmark&&&&&&&&\checkmark&&\checkmark&\checkmark&\checkmark&\checkmark&&&&&&&&&\checkmark&&\checkmark&&&&&\checkmark&&&&&&&&&&&&&&&&\checkmark&\checkmark&&&&&&\checkmark&\checkmark&\checkmark&&&\checkmark\\
Llama-3&&&&&&&&&&&&&&&&&&&&&&&&&&&&&&&&&&&&&&&&&&&&&&&&&&&&&&&&&&&&&&&\checkmark&\checkmark&&&\\
Mistral&&&&&&&&&&\checkmark&&&&&&&&&&\checkmark&\checkmark&\checkmark&\checkmark&&&&&&&&&&&&&&&&\checkmark&&&&&&&&&&&&&&&&&&&&&&&\checkmark&\checkmark&\checkmark&&&\\
Mixtral&&&&&&&&&&\checkmark&&&&&&&&&&&\checkmark&&&&&&&&&&&&&&&&&&\checkmark&&&&&&&&&&&&&&&&&&&&&&&&\checkmark&&&&\\
MPT&&&&&&&\checkmark&&&&&&&&&&&&&&&&&&&&&&&&&&&&&&&&&&&&&&&&&&&&&&&&&&&&&&&&&&&&\\
OLMo&&&&&&&&&&&&&&&&&&&&&&&&&&&&&&&&&&&&&&&&&&&&&&&&&&&&&&&&&&&&&&&\checkmark&&&&\\
OpenChat&&&&&&&&&&&&&&&&&&&&&\checkmark&&&&&&&&&&&&&&&&&&&&&&&&&&&&&&&&&&&&&&&&&&&&&&\\
OPT&&&&&&&&&&&&&&&&&&&&&&&&\checkmark&&&&&&&&&&&&&&\checkmark&&&&&&&&&&&&&&&&&&&&\checkmark&&&&\checkmark&&&&&\\
PaLM&&&&&&&&&&&&&&&&&&&&&&&&&&&&&&&&&\checkmark&\checkmark&&&&\checkmark&&&&&&&&&&&&&&&&&&&&&&&&&&&&&\\
Pythia&&&&&&&\checkmark&&&&&&&&&&&&&&&&&&&&&&&&&&&&&&&&&&&&&&&&&&&&&&&&&&&\checkmark&&&&&&&&&\\
Qwen&&&&&&&&&&&&&&&&&&&&&\checkmark&&&&&&&&&&&&&&&&&&&&&&&&&&&&&&&&&&&&&&&&&&&&&&\\
RedPajama&&&&&&&&&&&&&&&&&&&&&&&&&&&&&&&&&&&&&&&&&&&&&&&&&&&&&&&\checkmark&&&&&&&&&&&&\\
Solar&&&&&&&&&&&&&&&&&&&&&&&\checkmark&&&&&&&&&&&&&&&&&&&&&&&&&&&&&&&&&&&&&&&&&&&&\\
T5&&&&&&&&&&&&&&&&&&&&&&&&&&&&&&&&&&&&&&&&&&&&\checkmark&&&&&&&&&&&&&&\checkmark&&&&&&&&&\\
Tulu&&&&&&&&&&&&&&&&&&&&&&&&&&&&&&&&&&&&&&&&&&&&&&&&&&&&&&&&\checkmark&&&&&&&&&&&\\
Vicuna&&&&&\checkmark&&&&&&&&&&&&&&&&\checkmark&&\checkmark&&&&&&&&&&&&&&&&&&&&&&&&&&&&&&&&&&&&&&&&&&&&\\
Yi&&&&&&&&&&&&&&&&&&&&&\checkmark&&\checkmark&&&&&&&&&&&&&&&&&&&&&&&&&&&&&&&&&&&&&&&&&&&&\\
Zephyr&&&&&&&&&&&&&&&&&&&&\checkmark&\checkmark&&&&&&&&&&&&&&&&&&&&&&&&&&&&&&&&&&&&&&&&&&\checkmark&&&&

\end{tabular}
\caption{Overview of deployed models in surveyed works.}
\label{tab:models}
\end{table*}

\end{document}